\documentclass[letterpaper]{article}
\usepackage{natbib,alifeconf}
\usepackage{hyperref}
\pdfoutput=1

\usepackage{xcolor}
\hypersetup{
    colorlinks,
    linkcolor={red!50!black},
    citecolor={blue!50!black},
    urlcolor={blue!80!black}
}

\usepackage{amsthm}
\usepackage{amsmath,amssymb, amstext}
\usepackage{graphicx,caption,subcaption}
\usepackage{tikz}
\usetikzlibrary{arrows,fit,matrix,backgrounds,decorations.pathreplacing,calc}
\usetikzlibrary{positioning}
\usepackage{bm} 

\usepackage{xparse}

\bmdefine\bcolon{:}

\DeclareMathOperator{\mi}{mi}
\newcommand{\X}{\mathcal{X}}

\DeclareMathOperator{\pa}{pa}

\author{Martin Biehl$^{1}$, Takashi Ikegami$^{2}$ \and Daniel Polani$^1$ \\
\mbox{}\\
$^1$University of Hertfordshire  \\
$^2$University of Tokyo \\
m.biehl@herts.ac.uk}
\title{Towards information based spatiotemporal patterns as a foundation for \\agent representation in dynamical systems}
\date{\today}
\begin{document}
\maketitle

\begin{abstract}
We present some arguments why existing methods for representing agents fall short in applications crucial to artificial life. Using a thought experiment involving a fictitious dynamical systems model of the biosphere we argue that the metabolism, motility, and the concept of counterfactual variation should be compatible with any agent representation in dynamical systems. We then propose an information-theoretic notion of \emph{integrated spatiotemporal patterns} which we believe can serve as the basic building block of an agent definition. We argue that these patterns are capable of solving the problems mentioned before. We also test this in some preliminary experiments.
\end{abstract}

\section*{Introduction}
Within artificial life the concept of an \emph{agent} is fundamental. While studying life-as-it-could-be \citep{langton_artificial_1989}, we also study agents-as-they-could-be. An intuitive approach to agents is possibly to say that while not reproducing, i.e.\ during their individual lifetime, living organisms are agents. The concept of an agent in this way generalizes the concept of living organisms by de-emphasizing reproduction and with it Darwinian evolution. This point of view is also in line with the common practice of referring to robots or software programs as agents.

To give some more background \citep[see][for a more detailed discussion]{barandiaran_defining_2009}, there are a few properties that seem universally acknowledged as necessary for something to be referred to as an agent. The first of those is probably the capacity to act \citep{sep-agency}. However, \citet{barandiaran_defining_2009} notice that this already presupposes a form of individuality i.e.\ an ``entity'' that this capacity can be attributed to. Consequently they put the individuality criterion first. Having perception is another fairly uncontroversial requirement \citep[see e.g.][who for practical reasons ignore individuality and only require ``anything'' with perception and action]{russell_artificial_1995}. The last concept which is often alluded to is that of some form of goal-directedness of the agent. The goals agents should strive to achieve are usually required to be in the agents' own interest/intrinsic (e.g.\ preservation) and not the goals of some other agent (or programmer). For a thorough treatment on the latter point see \citet{froese_enactive_2009}. 

We broadly agree on the three (or four) main requirements of individuality, perception and action, as well as goal-directedness. However we are not satisfied with the lack of formal definitions of the notions themselves. We therefore take a different and particularly formal approach to the problem of defining agents. 

From the start we limit ourselves to a mathematically well-defined class of systems i.e.\ dynamical systems and their generalization to stochastic processes (we will refer to dynamical systems only, inclusion of stochastic processes is implied). We want to define agents as entities that can exist \emph{within} a dynamical system. In other words, we are looking for a \emph{representation} of agents within dynamical systems. While there is no guarantee that such a representation even exists, we believe that even if we fail, there might be some insights into \emph{why} we fail. This would also help the community to understand the concept of agents better. At the same time we expect that dynamical systems are actually a powerful enough class of systems to consider and that they will turn out to be able to contain convincing examples of agents. This optimism stems from the fact that dynamical systems have been extremely successful in modeling systems from physics through chemistry to biology. Compelling recent examples of dynamical systems which directly suggest they can contain agents can be found in \citet{virgo_thermodynamics_2011,bartlett_emergence_2015}. If we are successful, then we would obtain a definition of agents as \emph{features} of dynamical systems and eventually even of life as a feature of such systems. This would be a step towards defining life as a natural kind as required by \citet{cleland_defining_2002}. Finally our hope is to reveal the formal counterparts of the intuitions about living systems formulated by \citet{maturana_autopoiesis_1980}.

In order to make it more clear what we mean by agents within a dynamical system, consider the following example, to which we will come back throughout this paper. Say we had a dynamical system that is a sufficiently exact approximation of the entire biosphere including the influence of incoming (from the sun) and outgoing radiation. During individual runs of this dynamical system, given the right initial conditions, things should occur that correspond to living organisms in the real biosphere. In this case we would say that within this dynamical system agents occur. Our goal is to find a mathematical representation of these agents. Since agents are a generalization of living organisms, we expect that agent representations can at least in principle exhibit the full range of phenomena exhibited by living organisms. Limitations should only be due to the chosen dynamical system and not inherent to the agent representation.

This paper is a contribution to the discussion of the foundations of artificial life. It does not present a solution of how to represent agents in dynamical systems. Rather it defines a notion that can identify intrinsically distinguished spatiotemporal patterns that we believe can act as the basic building block on which a theory of agents can be built. The strategy we have in mind here is the following. First, define the spatiotemporal patterns which are suitable to represent both living (bacteria, animals, plants) and non-living (rocks, crystals) persistent objects. Then further classify those patterns into classes exhibiting features of agents such as perception, action and goal-directedness. Spatiotemporal patterns that satisfy all criteria will represent agents.

Also note that for the formal definition we here restrict ourselves to finite discrete-time distributed\footnote{\emph{Distributed} means that the state of the system is given as a set of values of multiple variables or degrees of freedom.} dynamical systems with an already given ``space-like'' and ``time-like'' structure. Examples of this include cellular automata. The restriction to finiteness is due to the improved clarity this choice brings with it. The notions we present are well-defined in various more general settings. However, currently the spatiotemporal-like structure seems necessary to us.

The rest of this paper is structured as follows. The next section will present three challenges to representations of agents in distributed dynamical systems. Then we look at the literature and discuss ways to represent agents formally and in how far they succeed or fail to meet our expectations. We will then quickly introduce the setting of distributed dynamical systems and formally introduce a notion that we believe is able to identify the spatiotemporal patterns. We give the intuition behind this notion and discuss it in the light of the three requirements mentioned before. Finally, we present some preliminary results in the setting of the game of life.

\section*{The problem of tracking agents}
As mentioned before we expect the agent representation to be able to deal with all features associated to living organisms in the biosphere. Two such features, their \emph{metabolism} and their \emph{motility} present a major challenge to the representation of agents. These two features both make it hard from a formal standpoint to ``keep track'' of the living organism within a trajectory of the system. A third feature, we call it \emph{counterfactual variation}, that we attribute to the biosphere makes it hard to represent agents reliably across different initial conditions. This list of three features makes no claim to be complete, obtaining a complete list is ongoing research however. The three features in more detail:
 \paragraph{Metabolism} All known living organisms are metabolic \citep{szathmary_evolutionary_2005} and the metabolism is also in the discussion for its possible role in the origins of life \citep[see e.g.][]{dyson_origins_1985,kauffman_investigations_2002}. This highlights its fundamental role and any final agent representation must accommodate for this. The difficulty is the following.
 
 Assume that the sufficiently accurate biosphere model from the introduction is particle-based, i.e.\ it describes the time evolution of the degrees of freedom of all the particles in the biosphere. Say at a time $t_1$ we are given all the particles (and their degrees of freedom) that pertain to some bacterium. Then a naive way to represent this bacterium would be to just track the time evolution of each of those particles. This we could (in principle) easily do in our model as well. However the particles that the bacterium is made up of at a later time $t_2$ are not the same as those at time $t_1$ because of the bacterium's metabolism. We would end up with particles floating around in the environment of the bacterium and not the bacterium itself. At the same time there would be particles that now pertain to the bacterium that we would not be tracking.
 
 An agent representation therefore would need to be able to track the bacterium itself and not just a specific set of degrees of freedom. One way this could be solved is by constantly readjusting or refocusing on the degrees of freedom pertaining to the agent.
 
 Note that we cannot be entirely sure that there is no coordinate transformation which would let us track living organisms (i.e.\ their corresponding structures in a model) by just following a particular set of degrees of freedom. However we are not aware of such a transformation. Any criterion however that can be used to refocus on an agent should be related to any coordinate transformation which results in the ``agents' own'' coordinate system.

 \paragraph{Motility} Living organisms can be motile and like the metabolism motility is in the discussion for its role in the origins of life \citep{froese_motility_2014}. A representation of agents must therefore be capable of dealing with motile agents. Motility plays a similar role for field theory models of the biosphere as the metabolism plays for particle based models. The degrees of freedom of a field theory are the field amplitudes at each point in space so that tracking those degrees of freedom over time only means to track the field in a specific region of space. However motility demands that agents are not bound to a fixed region in space. Then we again need to adjust (track) the degrees of freedom that constitute the agent as time passes.

\paragraph{Counterfactual variation} A third feature concerns another kind of variation of the degrees of freedom that can represent agents in a dynamical system. Namely variation under different initial conditions. We attribute to the biosphere a large variety of possible counterfactual histories that also support living organisms. Think of a biosphere where the continents are shifted a bit for example. This would not seem to necessarily destroy the possibility of the biosphere (geosphere) to contain living organism. 

Furthermore, we attribute to agents and living organisms the capability to behave differently under different environmental situations. The agent should be able to ``take a decision'' i.e. to walk either right or left, or eat the apple or the pear. Depending on these ``decisions'' the agent will again pertain to different degrees of freedom.

The counterfactual histories can be studied in the dynamical system setting by studying multiple trajectories through state space. Each trajectory corresponds to a different history (and possibly future). If the ``same'' agent occurs in two different trajectories it can behave differently in one from the other. This can be associated with different decisions \citep[e.g.][]{ikegami_uncertainty_1998}. 

The existence of benign counterfactual histories in our biosphere is an assumption and not possible to prove. However it is in line with the successful way physics models systems \citep[cf. the models of][]{virgo_thermodynamics_2011,bartlett_emergence_2015} and therefore in line with our general approach. 
Now given a set of counterfactual histories containing living organisms we expect that the degrees of freedom which in one history pertain to a living system at time $t$ need not pertain to a living system within another such history at $t$ (or in fact ever). More specifically, the degrees of freedom pertaining to a bacterium in one history need not pertain to any living organism in another. 

If the biosphere can contain living organisms within various counterfactual histories, then the dynamical systems model of the biosphere must be able to contain agents under various initial conditions. In that case the agent representation must be able to represent all the agents in all the trajectories where they occur. If for two different initial conditions the degrees of freedom pertaining to agents at time $t$ are different as well, then the agent representation must be able to exhibit this difference.  

\section*{Related work}
We should stress that we are only interested in work that relies on the \emph{intrinsic} properties of the dynamical systems itself to represent agents. References to concepts like action, perception, and goal-directedness, if they are not defined in terms of the dynamical system are not acceptable in this case.
The publication that most directly tackles the problem of agent representation that we are aware of is the insightful paper of \citet{krakauer_information_2014}. They solve the problems of metabolism and motility by evaluating informational measures of closure and autonomy of sets of random variables. Given a system represented by a set of random variables at each point in time (i.e.\ represented by a dynamical Bayesian network as also defined below) they propose an algorithm that decides whether to include a random variable at a specific time step into the set representing the agent or leave it in the set representing the environment. This decision is made according to whether the inclusion into the agent contributes to the closure or autonomy of the agent. What this approach lacks however is the capability to deal with counterfactual variation. Since they use measures like mutual information and mutual conditional information that average over all states of the random variables in order to decide whether they belong to the agent or not, the partition of the random variables at each time step is fixed for all possible trajectories of the system. In order to deal with counterfactual variation it must be possible to have one partition into agent and environment for one trajectory and another partition for another trajectory. The same argument remains true for any approach that results in a fixed partition of the nodes in a dynamical Bayesian network. This includes the work of \citet{balduzzi_detecting_2011} which results in a coarse-grained version of the network. The effective information that the glider contains about past states of the game of life, which was revealed in this work, should however be related the intrinsic spatiotemporal patterns that we investigate here.

Another very relevant and inspiring work is the work on the cognitive domain of the glider and autopoiesis in the game of life by \citet{beer_cognitive_2014,beer_characterizing_2014}. This approach  is capable of dealing with metabolism, motility, as well as counterfactual variation as it analyses spatiotemporal patterns and their internal mechanisms. The spatiotemporal patterns may have finite extension and can therefore occur or not occur within multiple trajectories at multiple times. The internal mechanisms are analyzed with respect to their production of the next \emph{spatial} pattern inside the spatiotemporal pattern. The only caveat seems to be that the analysis is quite time consuming and does not have formal expressions of all the involved notions. We use the notion of spatiotemporal patterns as presented by Beer and hope that the measure we propose contributes to the formalization of the notions in his work.

An approach that seems to solve the problem of metabolism and counterfactual variation is the Markov blanket-based clustering used by \citet{friston_life_2013}. As the interacting degrees of freedom vary over time in a particle based system, it is possible to define a time dependent adjacency (or interaction) matrix. From this matrix Friston derives a Markov blanket matrix which can be used to classify the degrees of freedom into hidden, sensory, active, and internal states. This nicely defines an agent like structure within the degrees of freedom and through the time dependence of the adjacency and therefore also the Markov blanket matrix allows for the degrees of freedom to vary within a single trajectory and across initial conditions. In the case of a field theoretical model where the adjacency of the degrees of freedom does not vary it is not directly obvious to us how to translate this. This means that motility could be a problem for the approach in such a model. However, it is definitely an alternative to our more information theory-based approach.

Methodologically, the framework of \citet{lizier_framework_2014} for distributed computation is very closely related to ours. They investigate localized versions of mutual information and conditional mutual information to track and highlight information transfer, storage, and modification in dynamical Bayesian networks. This reveals spatiotemporal patterns very similar to ours. The main formal difference is in fact that instead of localizing the (conditional) mutual information we localize multi-information in the same way. In this way our work is just a trivial extension of this work. The focus of our work however is different as we are not so much interested in phenomena that are related to computation and more interested in revealing spatiotemporal entities or objects which might form the basis of an agent definition. Related work on spatiotemporal filtering \citep{shalizi_automatic_2006,flecker_partial_2011} of cellular automata differs from ours in a similar way. While ``interesting'' phenomena in the time evolution of single trajectories are revealed, the focus is not on connecting the interesting phenomena together in order to obtain entities.

Conceptually our work is also closely related to the integrated information theory due to \citeauthor{tononi_measure_1994} Originally \citep{tononi_measure_1994} this involved measurements of multi-information whose localized (in the sense of \citeauthor{lizier_framework_2014}) version we also employ as an estimate of integration. Newer versions \citep{oizumi_phenomenology_2014,albantakis_intrinsic_2015} involve a more elaborate construction which, importantly, is also localized in a certain way. The latter detect distinguished integrated spatial patterns which are constructed to resolve ``what a system `is' from its own intrinsic perspective'' \citep{albantakis_intrinsic_2015}. How these spatial structures are connected in time however is not treated. Our approach also aims at revealing intrinsic structure but crucially looks for spatiotemporal patterns i.e.\ patterns with a temporal as well as a spatial extension or compositional structure. In \citet{tononi_information_2004,balduzzi_integrated_2008} temporal integration is mentioned with respect to optimal spatial and temporal scale or ``grain size'' detection. Our goal is different since we don't  want to find a coarse-graining here. We want to reveal the complete lifetimes of agents as a single spatiotemporal pattern. 

\section*{Dynamical Bayesian networks}
Finite discrete-time distributed dynamical systems and their stochastic counterparts can be represented by dynamical Bayesian networks. Dynamical here just means that there is an interpretation of time in those networks. Distributed means that, at each time step, there are multiple given random variables whose states together define the state of the entire network at that time step.

More formally, a (dynamical) Bayesian network is a directed acyclic graph $G=(V,E)$ with nodes $V$ and edges $E$. Each node $i$ has an associated random variable $X_i$ with state space $\X$ taking values $x_i \in \X$ (for simplicity we assume that all nodes have identical state spaces but this is not necessary for the definitions to hold). Furthermore each node is equipped with a mechanism $p_i(x_i|x_{\pa(i)})$ which gives the conditional probability distribution of $X_i$ given the parents $\pa(i)$ of node $i$ in $G$. Note that for any set $A \subseteq V$ we write $X_A: = (\{X_i|i \in A\})$ for the random variable composed of the random variables in $A$. We assume that our network has a set $V_0$ of nodes without parents. As \citet{ay_information_2008} note we can then define a partition of $V$ into $(V_0,V_1,V_2,...)$ (called \emph{time slices}) where $V_{t+1}:=\{i \in V \mid \exists j \in V_t, \pa(i)=j\}$. In general $\pa(V_{t+1}) \subseteq V_t$ since some nodes might not have any children. Here we assume $\pa(V_{t+1}) = V_t$. This allows us to interpret the various nodes in each $V_t$ as those nodes representing the state of the distributed system at time $t$. We can also interpret the cardinality of the set $V_t$ as the spatial extension of the state. In this paper this cardinality does not change over time just as e.g.\ in cellular automata. 

The defining property of Bayesian networks (including dynamic ones) is that the joint probability distribution\footnote{For a set of nodes $A$ we write $p_A$ for the probability distribution $p_A:\X^A \rightarrow [0,1]$.} $p_V$ can be factorized in a way compatible with the structure of the graph $G$ i.e.:
\begin{equation}
 p_V(x_V)=\prod_{i \in V} p_i(x_i|x_{\pa(i)}).
\end{equation} 

To relate the dynamical Bayesian network to dynamical systems note that the role of the dynamical law is played by the product of all mechanisms in $V_t$:
\begin{equation}
\label{eq:propagate}
 p_{V_{t+1}}(x_{V_{t+1}})=\sum_{x_{V_t}} \prod_{i \in V_{t+1}} p_i(x_i|x_{\pa(i)}) p_{V_t}(x_{V_t}).
\end{equation} 
Recall that $\bigcup_{i \in V_{t+1}} \pa(i) = V_t$ by definition. We can also write the above in terms of the Markov matrix:
\begin{equation}
 p(x_{V_{t+1}}|x_{V_t})= \prod_{i \in V_{t+1}} p_i(x_i|x_{\pa(i)}).
\end{equation}
In order to equip the dynamical Bayesian network with a join probability distribution $p_V$ we then only have to define an initial probability distribution $p_{V_0}$ and propagate it throughout the network according to Eq.\ \ref{eq:propagate}.

\section*{Trajectories and spatiotemporal patterns}
Here we formally define the notion of trajectories and spatiotemporal patterns. The class of the spatiotemporal patterns is very large and includes patterns that are of no specific interest. How we distinguish between those and more important patterns will be defined in the next sections.

A \emph{spatiotemporal pattern $x_O$ of a dynamical Bayesian network on graph $G=(V,E)$} is a set of nodes $O \subseteq V$ together a set of particular values $\{x_i \in \X|i \in O\}$.

A \emph{trajectory $x_V$ of a dynamical Bayesian network on graph $G=(V,E)$} is a spatiotemporal pattern with $p(x_V)>0$. In our setting this also means that there is an initial condition $x_{V_0}$ such that $x_V$ is possible under the time evolution induced by the Markov matrix or dynamical law.

We say that the spatiotemporal pattern \emph{$x_O$ occurs in a trajectory $x_V$} of the network iff $x_O \subseteq x_V$.

Employing the time slices $V_t$ of the network we can also look at the time slices $x_{O_t} := x_O \cap x_{V_t}$ of any spatiotemporal pattern $x_O$.

\section*{Integrated spatiotemporal patterns}
This section defines the notion of an integrated spatiotemporal pattern. Such patterns obey a condition which distinguishes them within the class of all spatiotemporal patterns. First we fix some further terminology. 

We define the \emph{the evidence for integration of an object $O$ with respect to a partition $\pi$ of $O$} as the local mutual information 
\begin{equation}
\label{eq:mi}
 \mi_\pi(x_O):= \begin{cases}
		  0 & \text{ if } p_O(x_O)=0, \\ 
                 \log \frac{p_O(x_O)}{\prod_{b_j \in \pi} p_{b_j}(x_{b_j})} & \text{ else}.
                \end{cases}
\end{equation} 
Then, we say a spatiotemporal pattern $x_O$ is \emph{integrated} iff for all possible partitions $\pi$ of the set $O$ of random variables the evidence for integration of $O$ with respect to $\pi$ is positive. Considering all possible partitions is also done by \citet{albantakis_intrinsic_2015}.

The interpretation of this is the following. The joint probability $p_O(x_O)$ is the probability that all the $x_i$ with $i \in O$ occur together within single trajectories. I.e. among all trajectories there are those in which $O$ occurs and their probability contributes to this joint probability. The probability $\prod_{b_j \in \pi} p_{b_j}(x_{b_j})$ however is the product of the probabilities that each part $x_{b_j}$ occurs by itself in any trajectory \emph{including} as part of $x_O$. If a part of $x_O$ often occurs by itself without the rest of $x_O$ occurring then this reduces the evidence for integration of $O$. This makes sense if we want to interpret the integrated spatiotemporal patterns as persistent objects like rocks, crystals, but also living organisms. If we consider for example a rock, the probability that a rock occurs at some point in time without a rock occurring at the previous and next time step in close vicinity is quite low, whereas the probability that where there was a rock before there will be a rock shortly after is quite high. In fact anytime that a spatial pattern (a time slice of a spatiotemporal pattern) causes (in an intuitive sense) another spatial pattern at the next time step their joint probability will rise and especially if the first spatial pattern is among the only causes of the second, their evidence for integration will be high. 

What about the spatial integration however? The existence of rock in one place probably does increase the probability for more rock to be around it, but not extremely. It is perfectly possible and occurs frequently, that the rock ends, also that it is just a small piece of rock. So the evidence for spatial integration might not be so strong. 

Now if we turn to living organisms, the evidence for temporal integration should also be high since they are autopoietic. Their spatial integration will probably be higher than that of rocks (and crystals) as half a bacterium is much less likely than a whole whereas half a rock is still a rock and those are not so uncommon. This reasoning scales up to larger living organisms.

We note that the evidence can also be interpreted in more information-theoretic terms. For example as the superfluous length of a codeword for the sequence $x_O$ when we base the encoding on the product probability distribution $\prod_{b_j \in \pi} p_{b_j}(x_{b_j})$ instead of on the joint probability. This will be discussed in more detail in future work. We also intend to investigate in how far the integrated spatiotemporal patterns are independent of (possibly moving) frames of reference. Since they are not only integrated across time slices but instead across any partition we are optimistic in this regard.

\section*{Integrated spatiotemporal patterns and the tracking problem}
The spatiotemporal patterns can solve the tracking problem. To see this take the perspective of time slices and say that at some time $t$ a living organism is a configuration of degrees of freedom which increases the probability of a particular configuration of other degrees of freedom at a subsequent time $t+\epsilon$ that is again a living organism. More specifically, a living organism will lack certain molecules before absorbing them, conversely there will be a surplus of other molecules before they are ejected from the living organism. Therefore the probability for molecular exchange will be higher than for maintaining the same composition. This means the spatiotemporal patterns traversing the degrees of freedom associated to the molecules will have a higher evidence for integration over time. Similarly, in the field-theoretic setting the field configuration represented by the spatial pattern will increase the probability of the neighboring degrees of freedom to assume a certain configuration. This leads to more evidence for the integration of the moving pattern.

With respect to the problem of counterfactual variation we can see the following. Integration is calculated directly for spatiotemporal patterns within a trajectory and the local mutual information vanishes for all spatiotemporal patterns that do not occur in this trajectory (see Eq.\ \ref{eq:mi}). Then, if the spatiotemporal patterns that occur in different trajectories are different, the integrated spatiotemporal patterns will also be different. Thus, if integrated spatiotemporal patterns represent agents, these can occur on some degrees of freedom in one trajectory and not occur on those in another. This means counterfactual variation won't be a problem.

\section*{Experimental indications}
We present here the results of three preliminary experiments. The first is conceived to hint at the kind of trajectories that show high evidence of integration. The second experiment suggests that traversal of degrees of freedom or motility/metabolism can at least in principle be detected by integration. Similarly the third experiment shows that in principle counterfactual variation is no obstacle for integration. 

All experiments use a $4 \times 4$ grid with game-of-life dynamics and toroidal boundary conditions as the distributed dynamical system. As the initial distribution $p_{V_0}$ we use the uniform distribution in order to explore the whole range of possible trajectories. We investigate only patterns covering three time steps $t=8,9,10$ and thereby neglect a lot of transient patterns that are difficult to interpret. In principle, however, our method does apply to transient patterns as well. Instead of integration we calculated only the \emph{evidence for integration with respect to the finest possible partition} (EVIFPP). The finest possible partition of a set $A$ of nodes is just the partition where each block is a set containing exactly one node in $A$. A positive EVIFPP is a necessary condition for integration and therefore a crude indication for it.

\begin{figure}[!ht]
 \centering
 \includegraphics[width=.95\linewidth]{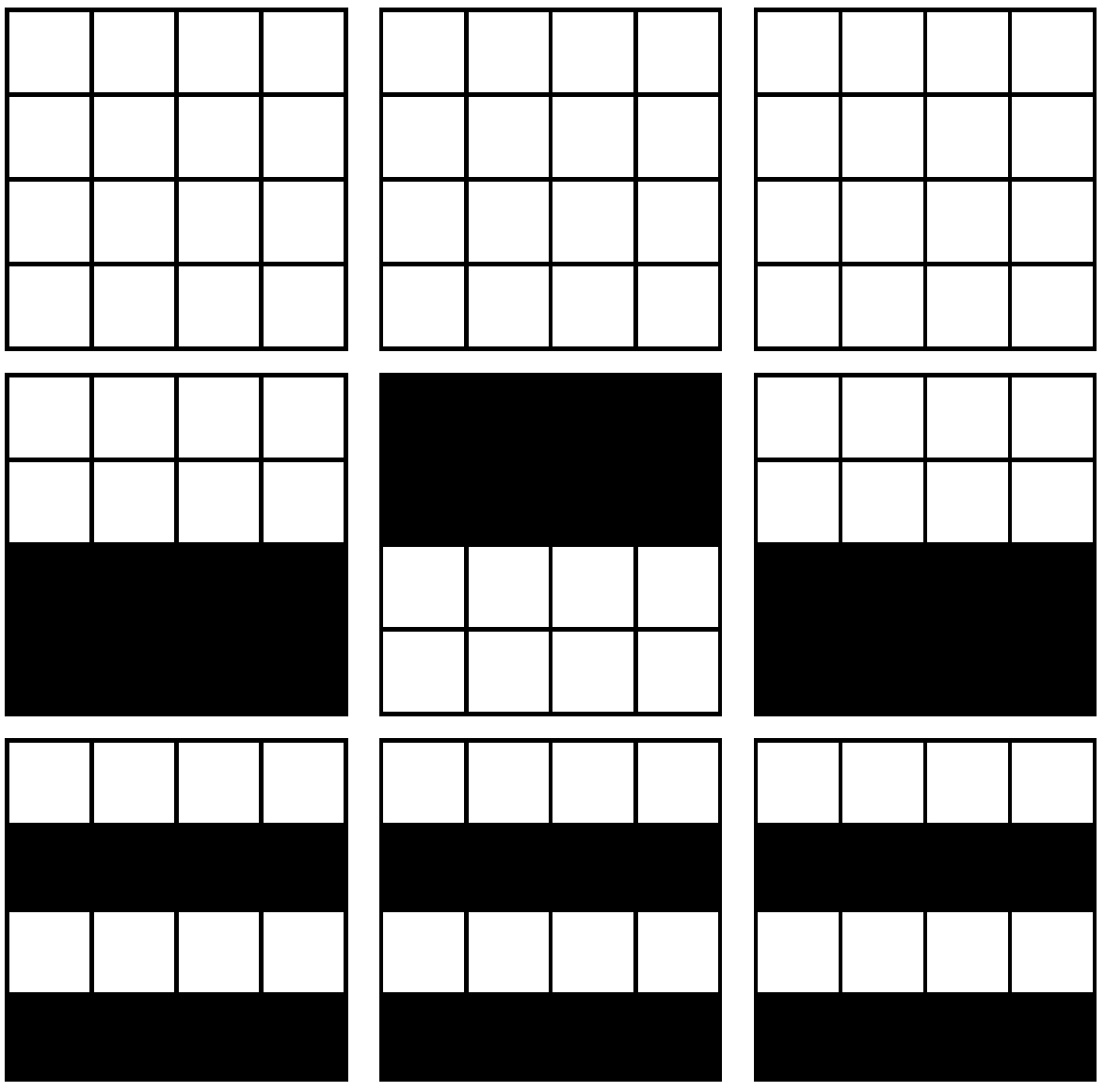}
 \caption{Three three-time step spatiotemporal patterns. Each row shows the three global spatial patterns that make up the spatiotemporal pattern $x_O$. The first row shows the blank spatiotemporal pattern and the others show the two patterns with the highest EVIFPP. The EVIFPP values are (from top to bottom) 4.9, 81.9, and 85.4 respectively.}
 \label{fig:tras}
\end{figure}

\begin{figure}[!ht]
 \centering
 \includegraphics[width=.95\linewidth]{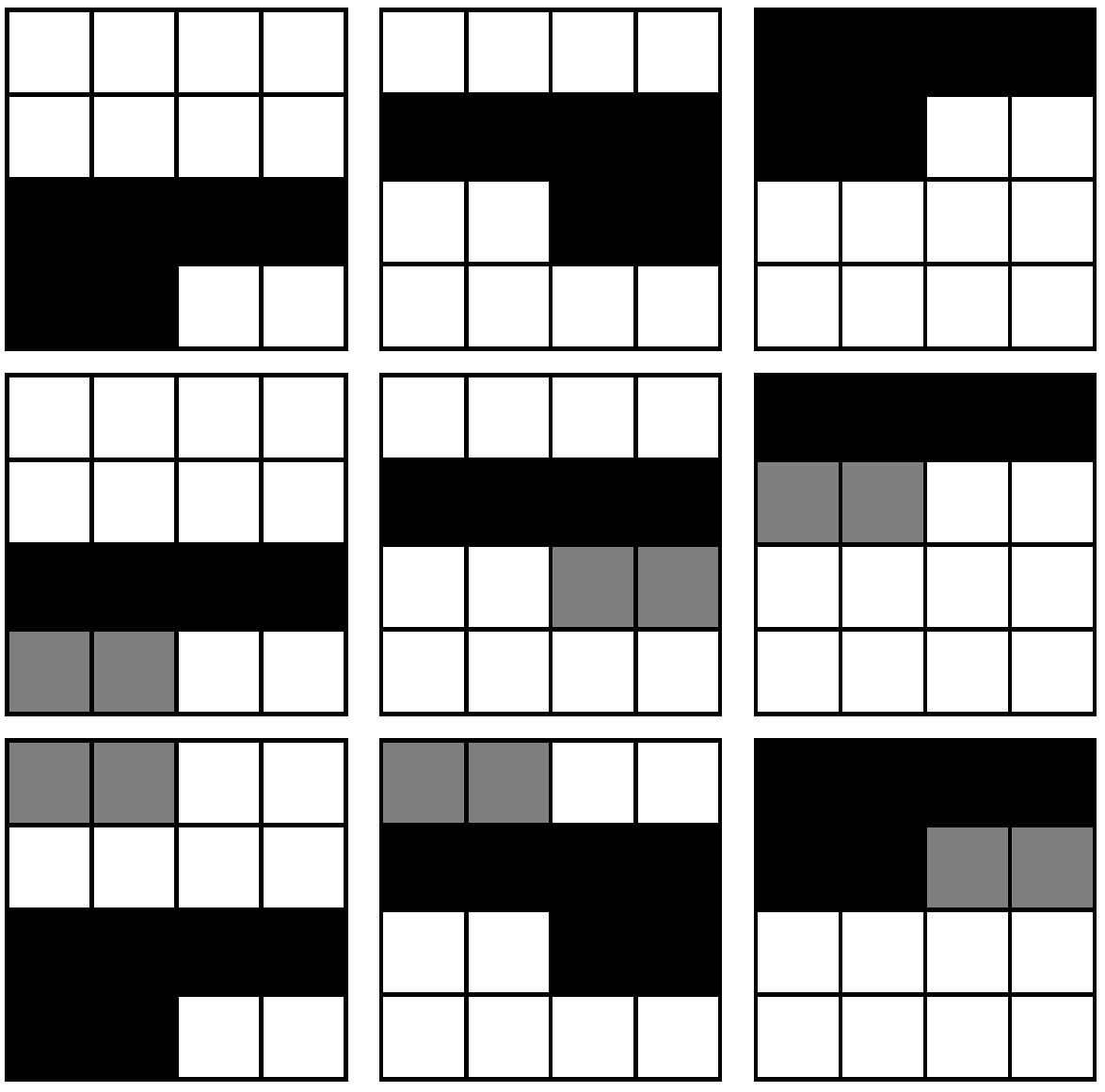}
 \caption{A three-time step part of a trajectory (can also be seen as a global spatiotemporal pattern) in the first row and two local spatiotemporal patterns on this trajectory in the second and third row. Both spatiotemporal patterns in rows two and three have $n=14$ specified cells per time slice. The second (third) row shows a pattern attaining the minimal (maximal) EVIFPP of 32.5 (54.4) among all patterns with $n=14$ on the trajectory of row one. The global spatiotemporal pattern of row one has EVIFPP of 55.0.}
 \label{fig:motility}
\end{figure}

For the first experiment we looked at all trajectories that differ at time steps $t=8,9,10$ (a lot of trajectories end up with all cells white at those times). For each of those trajectories we calculated the EVIFPP for the spatiotemporal pattern $x_O=(x_{V_8},x_{V_9},x_{V_{10}})$. So the time slices of $x_O$ are global states in this case. Since the $x_O$ are global states this is more an evaluation of the integration of the trajectories that result from the different initial conditions. In Fig.\ \ref{fig:tras} five such different global three-time-step patterns with high values of integration are shown including the completely blank spatiotemporal pattern and the spatiotemporal patterns (ignoring symmetric versions) with the highest EVIFPP. We can see that the blank spatiotemporal pattern has positive but much lower EVIFPP than some other patterns.
For the second experiment we chose a specific trajectory shown in the first row in Fig.\ \ref{fig:motility} which exhibits a moving pattern and searched through all patterns covering time steps $t=8,9,10$ and fixing $n=14$ cells (i.e.\ nodes of the dynamic Bayesian network) in each time slice $x_{O_t}$. We can see that the degrees of freedom (i.e. the cells or nodes) making up both the spatiotemporal pattern with minimal EVIFPP (second row in Fig.\ \ref{fig:motility}) as well as that with maximal EVIFPP (third row in Fig.\ \ref{fig:motility}) vary over the three time steps and adapt to the configuration of the global state. Note that the patterns with minimal and maximal EVIFPP are not unique.

\begin{figure}[!ht]
 \centering
 \includegraphics[width=.95\linewidth]{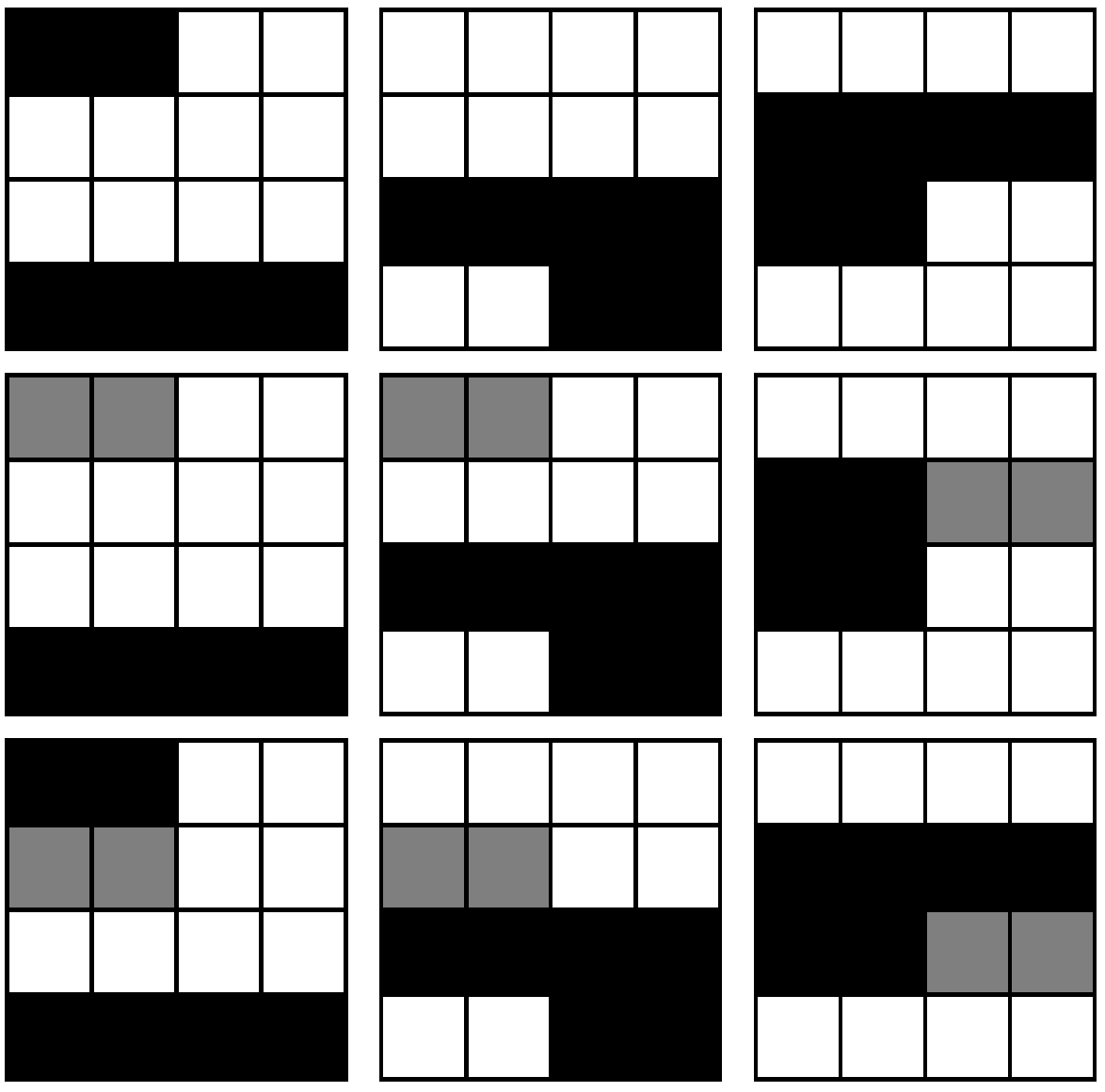}
 \caption{The three-time step part of the trajectory that results from shifting the values of the initial condition of the trajectory in the first row of Fig.\ \ref{fig:motility} ``down'' by one cell. The second row shows the spatiotemporal pattern with the same fixed nodes as the pattern that had maximal EVIFPP on the trajectory of Fig.\ \ref{fig:motility} but now on the shifted trajectory. The EVIFPP of this is 39.8. The third row shows the spatiotemporal pattern with the fixed nodes shifted ``down'' in the same way as the initial condition. This pattern has EVIFPP of 54.4 and is the maximal EVIFPP for patterns with $n=2$. As expected this is the same value we found for the pattern with the non-shifted nodes on the non-shifted trajectory.}
 \label{fig:counterfactual}
\end{figure}

For the third experiment changed the initial condition of the trajectory by shifting all values of the initial condition of the second experiment ``down'' one cell. This results in a different trajectory shown in the first row of Fig.\ \ref{fig:counterfactual}. We then evaluated the spatiotemporal pattern that results from fixing the same nodes as in the spatiotemporal pattern with maximal EVIFPP found in the second experiment on the changed trajectory (see row two in Fig.\ \ref{fig:counterfactual}). We also evaluated the EVIFPP of the spatiotemporal pattern that results from shifting the fixed nodes of maximal EVIFPP pattern in the same way as the initial condition (see row three in Fig.\ \ref{fig:counterfactual}). The pattern with the same fixed nodes as the pattern that formerly had maximal EVIFPP now has lower EVIFPP than the pattern with the nodes adapted to the new initial condition.

\section*{Discussion}
The first experiment shows that the completely blank trajectory has low spatiotemporal EVIFPP and that more ``interesting'' trajectories have higher EVIFPP (Fig.\ \ref{fig:tras}). This can also be done with other methods e.g.\ counting black cells. However, our method is general and doesn't use any prior knowledge e.g. which color of cells to count. For us this result is a necessary condition for further investigation.

The second experiment shows that the degrees of freedom pertaining to spatiotemporal patterns with high EVIFPP adapt over time to the  changing configurations of the system. This shows that EVIFPP is capable of solving the metabolism and motility problems. We expect that the same holds true for evidence of integration with respect to any partition and therefore also for integration itself.

The third experiment shows that under a variation of the initial condition the degrees of freedom pertaining to spatiotemporal patterns with high EVIFPP change accordingly. Since the different trajectories generated from changed initial conditions correspond to counterfactual histories this shows that the EVIFPP solves the problem of counterfactual variation. Again we expect this to carry over to integration. 

We note that larger grids become hard to evaluate computationally very fast. For square grids the size of the Markov matrix grows with $2^{a^2}$ where $a$ is the number of rows and columns of the grid. 
We also note that due to the very limited grid size we are studying any pair of cells is just separated by maximally one neighborhood cell. This leads to strong dependencies which might make it irrelevant to place unspecified cells around patterns like the blinker (as for example suggested by \citet{beer_characterizing_2014}). We had hoped to reveal such well known patterns and their extensions. Turning to larger grids is a next step in our research.  

\section*{Conclusion}
We have presented our current approach to representing agents in dynamical systems. Three criteria that we expect from such an agent representation were motivated with a thought experiment involving a dynamical systems model of the biosphere. The literature was reviewed in the light of these criteria. We also introduced our current candidate measure for identifying intrinsic spatiotemporal patterns in dynamical Bayesian networks. These patterns form the basic building blocks of our approach to representing agents. We argued that this approach can deal with the three criteria for agent representations that we have put forward. Experimentally we verified this for a crude approximation to our more involved concept of integration. However experimental results are currently inconclusive with respect to the tracking of structures that are actually relevant for agents. Therefore we see the value of this work mostly as a contribution to the discussion of the foundations of artificial life. Future work will bring more decisive results.

\section*{Acknowledgements}
We thank Christoph Salge, Olaf Witkowski, Nicola Catenacci-Volpi, Julien Hubert, Nathaniel Virgo, and Nicholas Guttenberg for discussions on this topic. Part of this research was performed during Martin Biehl's time as an International Research Fellow of the Japan Society for the Promotion of Science. The third author was supported in part by the H2020-641321 socSMCs FET Proactive project.

\bibliographystyle{apalike}
\bibliography{../bibliography.bib}
\end{document}